\documentclass[10pt,twocolumn,letterpaper]{article}

\usepackage{cvpr}      %

\usepackage{microtype}
\usepackage{graphicx}
\usepackage{booktabs} %
\usepackage{url}            %
\usepackage{nicefrac}       %
\usepackage[dvipsnames]{xcolor}         %
\usepackage[export]{adjustbox}

\usepackage{subcaption}
\usepackage{scalerel}
\usepackage[normalem]{ulem}
\usepackage{wrapfig}
\usepackage{siunitx}
\usepackage{tabularx}
\usepackage{mwe}%
\usepackage[outline]{contour}
    \contourlength{0.1em}

\usepackage{tikz}
\usepackage{pgfplots}
\usepgfplotslibrary{groupplots}
\usepgfplotslibrary{colorbrewer}
\pgfplotsset{colormap={correlation_cm}{rgb255=(234,52,35) rgb255=(242,240,240) rgb255=(7,0,245)}}
\usetikzlibrary{fit}

\usepackage{listofitems} %
\usetikzlibrary{arrows.meta} %
\tikzset{>=latex} %

\RequirePackage{xspace}

\usepackage{amsmath}
\usepackage{amssymb}
\usepackage{mathtools}
\usepackage{amsthm}

\theoremstyle{plain}
\newtheorem{theorem}{Theorem}[section]

\theoremstyle{definition}
\newtheorem{definition}[theorem]{Definition}

\theoremstyle{remark}

\newcommand{\myparagraph}[1]{\vspace{0.25em}\noindent\textbf{#1}\hspace{0.5em}}
\newcommand{\myparagraphnospace}[1]{\noindent\textbf{#1}}

\definecolor{colorConceptOne}{RGB}{234,0,234}
\definecolor{colorConceptTwo}{RGB}{0,148,148}

\usepackage[accsupp]{axessibility}

\definecolor{cvprblue}{rgb}{0.21,0.49,0.74}
\usepackage[pagebackref,breaklinks,colorlinks,allcolors=cvprblue]{hyperref}

\usepackage[capitalize]{cleveref}
\crefname{section}{Sec.}{Secs.}
\Crefname{section}{Section}{Sections}
\Crefname{table}{Table}{Tables}
\crefname{table}{Tab.}{Tabs.}
\Crefname{figure}{Figure}{Figures}
\crefname{figure}{Fig.}{Figs.}

\title{Disentangling Polysemantic Channels in Convolutional Neural Networks}

\newcommand{\authorstep}{\hspace{0.5cm}}
\newcommand{\affiliationstep}{\hspace{0.5cm}}
\author{Robin Hesse\textsuperscript{1}
\authorstep Jonas Fischer\textsuperscript{2} \authorstep
Simone Schaub-Meyer\textsuperscript{\normalfont{}1,3}
\authorstep Stefan Roth\textsuperscript{\normalfont{}1,3}\\[4pt]
\normalsize{\textsuperscript{1}Technical University of Darmstadt\affiliationstep \textsuperscript{2}Max Planck Institute for Informatics \affiliationstep \textsuperscript{3}hessian.AI
}}

\usepackage{fancyhdr}
\usepackage{setspace}

\fancypagestyle{firststyle}
{

	\fancyhf{}
	\lfoot{{\footnotesize\begin{spacing}{.5}\parbox{\linewidth}{\vspace{2.5em}%
					To appear in Proceedings of the \emph{IEEE/CVF Conference on Computer Vision and Pattern Recognition Workshops (CVPRW)}, The First Workshop on Mechanistic Interpretability for Vision (MIV), 2025.%
					\\\hrule\vspace{\baselineskip}
					\copyright~2025 IEEE. Personal use of this material is permitted. Permission from IEEE must be obtained for all other uses, in any current or future media, including reprinting/republishing this material for advertising or promotional purposes, creating new collective works, for resale or redistribution to servers or lists, or reuse of any copyrighted component of this work in other works. 
	}\end{spacing}}}
}

\begin{document}

\twocolumn[{%
\renewcommand\twocolumn[1][]{#1}%
\maketitle
\begin{center}
\begin{adjustbox}{minipage=\linewidth, scale=.95}
    \centering
    \vspace{-1.9em}
    \captionsetup{type=figure}
        \captionsetup[subfigure]{labelformat=empty}
    \centering
        \begin{subfigure}{0.32\linewidth}
           \centering
           \begin{tikzpicture}

    \def\xdistance{1.6};

    \def\xdistanceoffset{0.4};
    \def\ydistance{1.1};
    \def\mywidth{0.15};

    \node[anchor=center,inner sep=0] (frame_center) at (0,0)        
    {\includegraphics[width=0.65\textwidth]{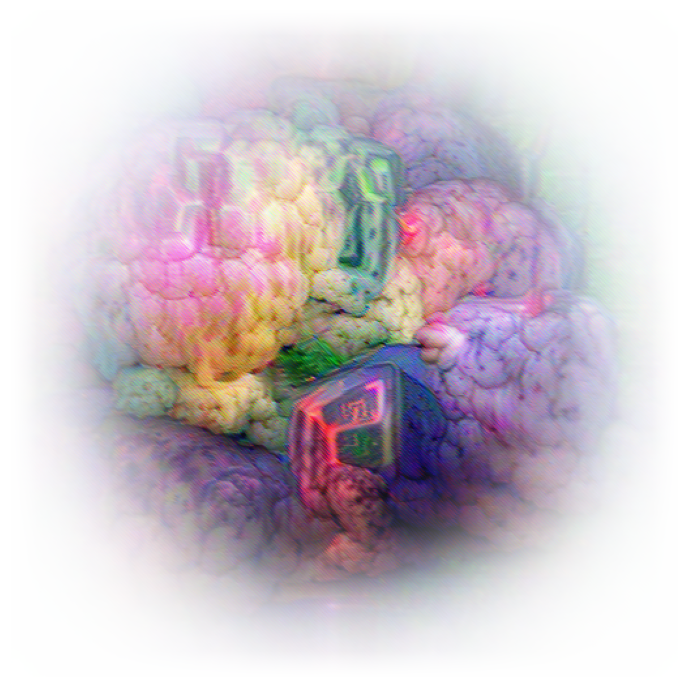}};

    \node[anchor=center,inner sep=0pt,clip,rounded corners=0.2cm] (frame0) at (-\xdistance + \xdistanceoffset,\ydistance)        
    {\includegraphics[width=\mywidth\textwidth]{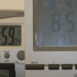}};
    \node[draw=colorConceptOne,very thick,fit=(frame0),rounded corners=.2cm,inner sep=0pt] {};

    \node[anchor=center,inner sep=0pt,clip,rounded corners=0.2cm] (frame1) at (-\xdistance,0)        
    {\includegraphics[width=\mywidth\textwidth]{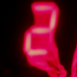}};
    \node[draw=colorConceptOne,very thick,fit=(frame1),rounded corners=.2cm,inner sep=0pt] {};

    \node[anchor=center,inner sep=0pt,clip,rounded corners=0.2cm] (frame2) at (-\xdistance + \xdistanceoffset,-\ydistance)        
    {\includegraphics[width=\mywidth\textwidth]{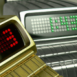}};
    \node[draw=colorConceptOne,very thick,fit=(frame2),rounded corners=.2cm,inner sep=0pt] {};

    \node[anchor=center,inner sep=0pt,clip,rounded corners=0.2cm] (frame3) at (\xdistance - \xdistanceoffset,\ydistance)        
    {\includegraphics[width=\mywidth\textwidth]{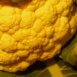}};
    \node[draw=colorConceptTwo,very thick,fit=(frame3),rounded corners=.2cm,inner sep=0pt] {};

    \node[anchor=center,inner sep=0pt,clip,rounded corners=0.2cm] (frame4) at (\xdistance,0)        
    {\includegraphics[width=\mywidth\textwidth]{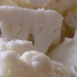}};
    \node[draw=colorConceptTwo,very thick,fit=(frame4),rounded corners=.2cm,inner sep=0pt] {};

    \node[anchor=center,inner sep=0pt,clip,rounded corners=0.2cm] (frame5) at (\xdistance - \xdistanceoffset,-\ydistance)        
    {\includegraphics[width=\mywidth\textwidth]{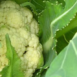}};
    \node[draw=colorConceptTwo,very thick,fit=(frame5),rounded corners=.2cm,inner sep=0pt] {};

\end{tikzpicture}

           \vspace{-0.65em}
        \caption{Original}
        \end{subfigure}\hspace{1mm}
        \begin{subfigure}{0.32\linewidth}
            \centering
            \begin{tikzpicture}

    \def\xdistance{1.6};

    \def\xdistanceoffset{0.4};
    \def\ydistance{1.1};
    \def\mywidth{0.15};

    \node[anchor=center,inner sep=0] (frame_center) at (0,0)        
    {\includegraphics[width=0.65\textwidth]{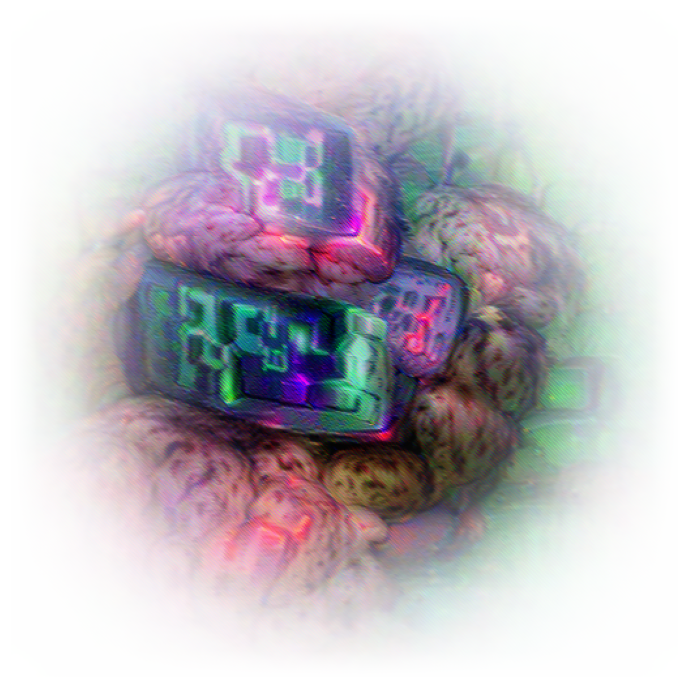}};

    \node[anchor=center,inner sep=0pt,clip,rounded corners=0.2cm] (frame0) at (-\xdistance + \xdistanceoffset,\ydistance)        
    {\includegraphics[width=\mywidth\textwidth]{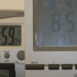}};
    \node[draw=colorConceptOne,very thick,fit=(frame0),rounded corners=.2cm,inner sep=0pt] {};

    \node[anchor=center,inner sep=0pt,clip,rounded corners=0.2cm] (frame1) at (-\xdistance,0)        
    {\includegraphics[width=\mywidth\textwidth]{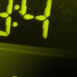}};
    \node[draw=colorConceptOne,very thick,fit=(frame1),rounded corners=.2cm,inner sep=0pt] {};

    \node[anchor=center,inner sep=0pt,clip,rounded corners=0.2cm] (frame2) at (-\xdistance + \xdistanceoffset,-\ydistance)        
    {\includegraphics[width=\mywidth\textwidth]{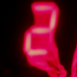}};
    \node[draw=colorConceptOne,very thick,fit=(frame2),rounded corners=.2cm,inner sep=0pt] {};

    \node[anchor=center,inner sep=0pt,clip,rounded corners=0.2cm] (frame3) at (\xdistance - \xdistanceoffset,\ydistance)        
    {\includegraphics[width=\mywidth\textwidth]{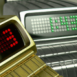}};
    \node[draw=colorConceptOne,very thick,fit=(frame3),rounded corners=.2cm,inner sep=0pt] {};

    \node[anchor=center,inner sep=0pt,clip,rounded corners=0.2cm] (frame4) at (\xdistance,0)        
    {\includegraphics[width=\mywidth\textwidth]{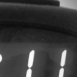}};
    \node[draw=colorConceptOne,very thick,fit=(frame4),rounded corners=.2cm,inner sep=0pt] {};

    \node[anchor=center,inner sep=0pt,clip,rounded corners=0.2cm] (frame5) at (\xdistance - \xdistanceoffset,-\ydistance)        
    {\includegraphics[width=\mywidth\textwidth]{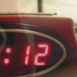}};
    \node[draw=colorConceptOne,very thick,fit=(frame5),rounded corners=.2cm,inner sep=0pt] {};

\end{tikzpicture}

            \vspace{-0.65em}
        \caption{Disentangled 1}
        \end{subfigure}\hspace{1mm}
        \begin{subfigure}{0.32\linewidth}
           \centering
           \begin{tikzpicture}

    \def\xdistance{1.6};

    \def\xdistanceoffset{0.4};
    \def\ydistance{1.1};
    \def\mywidth{0.15};
    
    \node[anchor=center,inner sep=0] (frame_center) at (0,0)        
    {\includegraphics[width=0.65\textwidth]{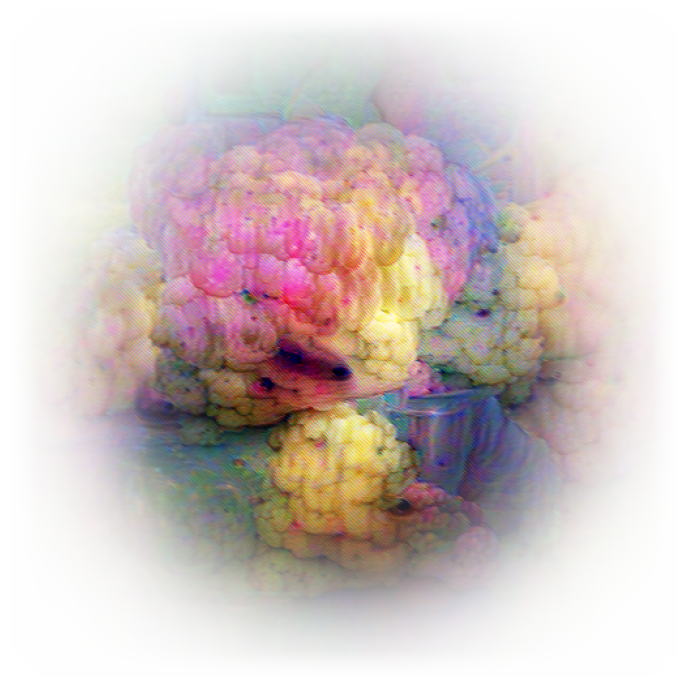}};

    \node[anchor=center,inner sep=0pt,clip,rounded corners=0.2cm] (frame0) at (-\xdistance + \xdistanceoffset,\ydistance)        
    {\includegraphics[width=\mywidth\textwidth]{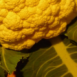}};
    \node[draw=colorConceptTwo,very thick,fit=(frame0),rounded corners=.2cm,inner sep=0pt] {};

    \node[anchor=center,inner sep=0pt,clip,rounded corners=0.2cm] (frame1) at (-\xdistance,0)        
    {\includegraphics[width=\mywidth\textwidth]{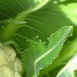}};
    \node[draw=colorConceptTwo,very thick,fit=(frame1),rounded corners=.2cm,inner sep=0pt] {};

    \node[anchor=center,inner sep=0pt,clip,rounded corners=0.2cm] (frame2) at (-\xdistance + \xdistanceoffset,-\ydistance)        
    {\includegraphics[width=\mywidth\textwidth]{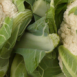}};
    \node[draw=colorConceptTwo,very thick,fit=(frame2),rounded corners=.2cm,inner sep=0pt] {};

    \node[anchor=center,inner sep=0pt,clip,rounded corners=0.2cm] (frame3) at (\xdistance - \xdistanceoffset,\ydistance)        
    {\includegraphics[width=\mywidth\textwidth]{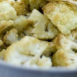}};
    \node[draw=colorConceptTwo,very thick,fit=(frame3),rounded corners=.2cm,inner sep=0pt] {};

    \node[anchor=center,inner sep=0pt,clip,rounded corners=0.2cm] (frame4) at (\xdistance,0)        
    {\includegraphics[width=\mywidth\textwidth]{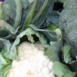}};
    \node[draw=colorConceptTwo,very thick,fit=(frame4),rounded corners=.2cm,inner sep=0pt] {};

    \node[anchor=center,inner sep=0pt,clip,rounded corners=0.2cm] (frame5) at (\xdistance - \xdistanceoffset,-\ydistance)        
    {\includegraphics[width=\mywidth\textwidth]{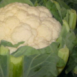}};
    \node[draw=colorConceptTwo,very thick,fit=(frame5),rounded corners=.2cm,inner sep=0pt] {};

\end{tikzpicture}

           \vspace{-0.65em}
        \caption{Disentangled 2}
        \end{subfigure}
    \vspace{-0.5em}     
    \caption{\textit{Strongly activating image patches and MACO~\cite{Fel:2023:UFV} visualizations for channel $\#1660$ of an ImageNet-trained ResNet-50 and the two corresponding disentangled channels (see \cref{subsec:disentangle}).} The original channel $\#1660$ is polysemantic as it strongly activates for unrelated concepts occurring in ``digital clock'' and ``cauliflower'' images. %
    After disentangling, we obtain channels that are significantly stronger activating \textit{either} to concepts occurring in ``digital clock'' (Disentangled 1) \textit{or} to concepts occurring in ``cauliflower'' (Disentangled 2).
    \vspace{-0.1em}
    }
    \label{fig:visualization_handcrafted}

\end{adjustbox}
\end{center}%
\smallskip
}]

\begin{abstract}
  
Mechanistic interpretability is concerned with analyzing individual components in a (convolutional) neural network (CNN) and how they form larger circuits representing decision mechanisms. 
These investigations are challenging since CNNs frequently learn polysemantic channels that encode distinct concepts, making them hard to interpret.
To address this, we propose an algorithm to disentangle a specific kind of polysemantic channel into multiple channels, each responding to a single concept. 
Our approach restructures weights in a CNN, utilizing that different concepts within the same channel exhibit distinct activation patterns in the previous layer.
By disentangling these polysemantic features, we enhance the interpretability of CNNs, ultimately improving explanatory techniques such as feature visualizations.\footnote{Code available at \href{https://github.com/visinf/disentangle-channels}{github.com/visinf/disentangle-channels}.}

\end{abstract}

\thispagestyle{firststyle}
\section{Introduction}
\label{sec:introduction}

Although convolutional neural networks (CNNs) are vital to many areas of computer vision, a major limitation is their increasing complexity, impeding their interpretability. 
As a result, the idea of \textit{mechanistic interpretability} has been proposed,
which focuses on understanding the information encoded in individual neurons or channels and how these form larger circuits that encode the decision mechanism of the network.
In particular, feature visualization by optimization~\cite{Erhan:2009:VHL}, \textit{i.e.}, generating an image that maximizes the activation of a CNN neuron/channel, is a widely-used method for visualizing encoded concepts. A common issue for such approaches is the presence of \textit{polysemantic} neurons, resp.\ channels, that activate for multiple unrelated concepts. Those are less interpretable because post-hoc explanation methods like feature visualizations have not been designed to visualize multiple concepts simultaneously.
Polysemanticity can occur because the CNN uses a non-privileged basis to encode concepts~\cite{Elhage:2022:TMS} or because the CNN learns to encode more concepts than channel dimensions by superimposing the concepts in the channel~\cite{Elhage:2022:TMS}. %
To resolve this issue,
there have been efforts to understand the functioning of individual, polysemantic neurons~\cite{Mu:2020:CEN, Nguyen:2016:MFV, Oikarinen:2024:LEI, Rosa:2023:TFU} or to identify \textit{monosemantic} directions in feature space~\cite{Kim:2018:IBF,OMahony:2023:DNR}. 
We take a step further and aim to \textit{disentangle} polysemantic channels.
This allows moving back to analyzing \textit{monosemantic} concepts learned by \textit{individual} channels, which is more in line with the principles of mechanistic interpretability.
As opposed to concurrent work~\cite{Dreyer:2024:TPN}, we propose an \textit{explicit} disentanglement instead of a ``virtual'' one; %
our approach hence enables, \textit{inter alia}, 
feature visualization through optimization (see \cref{fig:visualization_handcrafted}). %
Specifically, our main contributions can be summarized as:
\textit{(1)} We propose a novel approach to identify polysemantic feature channels in CNNs.
\textit{(2)} Using the identified channels, we introduce a method to disentangle a polysemantic channel into multiple channels, only activating for one of the concepts each.

\section{Related Work}
\label{sec:related_work}

\begin{figure*}[t]
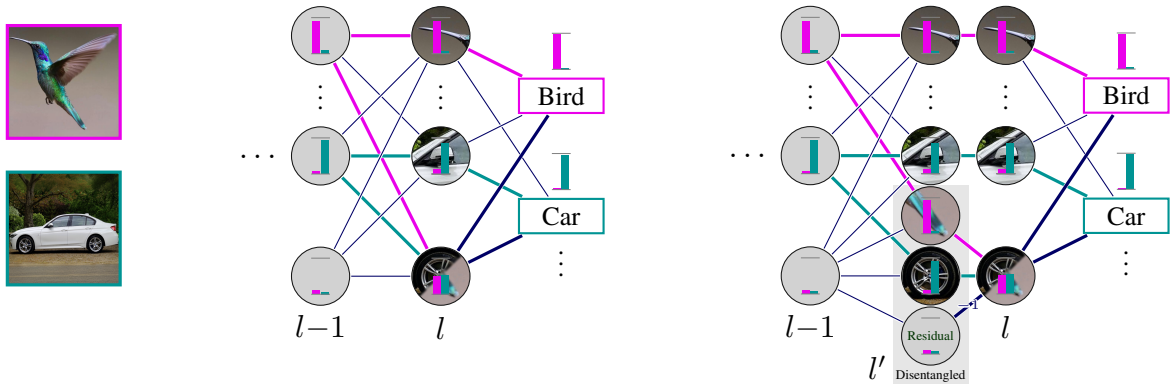

    \vspace{-4.5em}
    \captionsetup[subfigure]{labelformat=empty}
    \centering
        \begin{subfigure}{0.1\linewidth}
           \centering
           \input{figures/nn_illustration/input_images}
        \end{subfigure}\hspace{5mm}
        \begin{subfigure}{0.37\linewidth}
           \centering
           \input{figures/nn_illustration/nn_original}
        \end{subfigure}\hspace{2mm}
        \begin{subfigure}{0.4\linewidth}
           \centering
           \input{figures/nn_illustration/nn_disentangled}
        \end{subfigure}
    \vspace{-1.5em}
    \caption{\textit{Illustration of our proposed disentanglement approach.} We show two input samples from two different classes \emph{(left)}, the original neural network \emph{(middle)}, and our disentangled neural network \emph{(right)}. The bar plots in each channel indicate how active that channel is for each of the two input images. The color of the edge indicates if that edge is mainly propagating information from the first image, from the second image, or from both/none. The images in the nodes represent the encoded concepts.%
    }
    \label{fig:neural_network}
    \vspace{-0.5em}
\end{figure*}

Closest to our work is \cite{Dreyer:2024:TPN}, which disentangles polysemantic neurons by assigning their active circuit to specific circuit clusters corresponding to each concept. While this ``virtual'' disentanglement enables ``classifying'' the active concept, it is not inherently part of the network, making standard XAI methods, such as feature visualization, non-trivial or even impossible. In contrast, we propose an \textit{explicit} disentanglement (see \cref{fig:neural_network}) that allows direct application of established mechanistic interpretability tools.
Another line of research visualizes the \textit{different} concepts encoded in a \textit{single} feature~\cite{Mu:2020:CEN, Nguyen:2016:MFV, Oikarinen:2024:LEI, Rosa:2023:TFU}. \Eg, %
\cite{Nguyen:2016:MFV} proposes an initialization scheme, allowing for the feature visualization of different concepts encoded in a polysemantic feature. Instead of focusing on individual features, \cite{Kim:2018:IBF,OMahony:2023:DNR} aims to find meaningful \textit{directions in feature space} that correspond to a \textit{monosemantic} concept. %
Accordingly, a concept activation vector~\cite{Kim:2018:IBF} is defined as the normal to the classification boundary in the feature space of the layer of interest, separating images that contain a concept of interest and images without that concept. 
Also closely related to our approach are concept whitening~\cite{Chen:2020:CWI} and sparse autoencoders~\cite{Cunningham:2023:SAF}. %
In the latter, a layer of interest is projected into a higher dimensional space while enforcing sparsity constraints to obtain a more monosemantic encoding. %
However, this approach is only an approximation of the original network. Further, it does not allow for an understanding of \textit{individual} neurons and comes with its own set of challenges~\cite{heap2025sparseautoencodersinterpretrandomly, paulo2025sparseautoencoderstraineddata, wu2025axbenchsteeringllmssimple}.

\section{Disentangling
Polysemantic Channels}
\label{sec:method}

\newcommand{\channel}{c}
\newcommand{\class}{t}
\newcommand{\model}{f}
\newcommand{\layer}{{l}}
\newcommand{\layerbefore}{{l^{-1}}}
\newcommand{\layerwosub}{l}
\newcommand{\attribution}{\mathcal{A}}
\newcommand{\threshold}{\tau}
\newcommand{\thresholdARV}{\gamma}

We consider a classification setup with a set of images $X$ belonging to one of $n$ classes $\class \in \{1\ldots n\}$. 
We let $f_{l_1}(x)_{i_1}$ denote the channel $i_1$ in the output of a CNN $f$ at layer $l_1$ on input $x$. The relevance/attribution of an earlier channel $i_2$ in layer $l_2$ to the activation of a later channel $i_1$ in layer $l_1, \; l_1 > l_2$ can be computed via
\begin{equation}
\attribution_{l_2,i_2}^{l_1,i_1}(x; f) = \operatorname{sum} \left( \model_{\layerwosub_2}(x)_{i_2} \, \frac{\partial \, \operatorname{sum}\big( \model_{l_1}(x)_{i_1} \big)}{\partial \model_{\layerwosub_2}(x)_{i_2}} \right)\,,
\end{equation}
where $\operatorname{sum}(\cdot)$ denotes the sum along the spatial dimensions, aggregating a two-dimensional channel into a scalar (similar to global average pooling). Due to its simplicity, we here use Input$\times$Gradient~\cite{Shrikumar:2017:LIF} to define $\attribution$, which we found to work sufficiently well. However, $\attribution$ could also rely on other attribution methods, such as %
LRP~\cite{Bach:2015:PWE} or %
IG~\cite{Sundararajan:2017:AAD}. %

\subsection{Identifying polysemantic channels}\label{subsec:identify}

\paragraph{Finding candidate channels.} %
We assume that a single channel is detecting one or more (abstract) concepts.\footnote{According to the principles of superposition and non-privileged bases~\cite{Elhage:2022:TMS}, some concepts are encoded by a combination of channels. However, in the general sense, each channel encodes some information, which could also be regarded as a concept.}
We consider channels as \textit{potentially} polysemantic if the concepts they encode are occurring in different classes, \ie, if they are relevant to multiple classes.
For each class index $\class$ and the corresponding training images $X^{(\class)}$, we compute the relative attribution of the logit target output with respect to our layer of interest $l$ and consider a channel as relevant if its relative attribution exceeds a threshold $\threshold$ for a fraction $p$ of $X^{(\class)}$. %
Formally, a channel $\channel$ in layer $l$ is relevant for the $\class^\text{th}$ class if 
\begin{equation}\label{eq:relevant_channel}
\left(\frac{1}{|X^{(\class)}|} \sum_{x \in X^{(\class)}}\left[ \frac{\attribution_{l,\channel}^{l^\omega, \class}(x; f)}{\sum_{i \in \mathcal{C}_l} \attribution_{l,i}^{l^\omega, \class}(x; f)} > \threshold \right]\right) > p\,,
\end{equation}
with $[\cdot]$ denoting the Iverson bracket~\cite{Knuth:1992:TNN}, $l^\omega$ denoting the final layer, and $\mathcal{C}_l$ all channel indices in layer $l$.
Channels relevant for two or more classes are considered \textit{polysemanticity candidates}. This can happen because \textit{(1)} a channel activates for a single concept shared by classes (\textit{e.g.}, different dog species), meaning no polysemanticity occurs or \textit{(2)} a channel activates for multiple concepts, resulting in polysemanticity. %

\myparagraph{Identifying polysemanticity in candidate channels} requires to distinguish between possibility \textit{(1)} and \textit{(2)}. We build our approach on two observations: %
similar concepts should have similar activations not only in our layer of interest $l$, but also in the preceding layer $l{-}1$; and polysemanticity can be caused by superposition, which preferably occurs in sparse configurations, \textit{i.e.}, %
when superimposed concepts -- \textit{e.g.}, a beak and a car mirror -- rarely co-occur. 
From this, we infer that scenario \textit{(1)} (no polysemanticity) occurs predominantly for a candidate channel relevant for classes $\class_1$ and $\class_2$ if the channel activations in the preceding layer $l{-}1$ relevant for $\channel$ are \textit{similar} for images from $\class_1$ and $\class_2$. Vice versa, scenario \textit{(2)} (polysemanticity) occurs predominantly if relevant channel activations in $l{-}1$ are \textit{different} for images from $\class_1$ and $\class_2$. 
To distinguish the two, we first compute the average relevance vectors ($\text{ARV} \in \mathbb{R}^{|\mathcal{C}_{l-1}|}$) in layer $l{-}1$ %
\begin{gather}
\begin{aligned}
\text{ARV}(l, \channel, \class; f) =   \left( \frac{1}{|X^{(\class)}|} \sum_{x \in X^{(\class)}}  \attribution_{l{-}1,i}^{l,\channel}(x; f) \right)_{i \in \mathcal{C}_{l-1}} ,
\end{aligned}
\raisetag{30pt}
\end{gather}
with $\mathcal{C}_{l-1}$ being all channel indices in $l{-}1$. The ARV indicates the average relevance of each channel in layer $l{-}1$ to channel $c$ in layer $l$ for images of class $t$. With $S_{\operatorname{cos}}$ denoting the cosine similarity, we get the following definition.%

\begin{definition}[$\gamma$\nobreakdash-polysemanticity]\label{def:superposition}
Given a channel $\channel$ in layer $l$ that was found to be relevant for target classes $\class_1, \class_2$ as per Ineq.~(\ref{eq:relevant_channel}), we say that $\channel$ is affected by $\gamma$\nobreakdash-polysemanticity for the concepts it detects in $\class_1$ and $\class_2$ \wrt a threshold $\thresholdARV$ iff 
\begin{equation}
S_{\operatorname{cos}}\big(\text{ARV}(l, \channel, \class_1; f), \text{ARV}(l, \channel, \class_2; f)\big) < \thresholdARV.
\end{equation}
\end{definition}
\noindent In the following, ``polysemanticity'' will refer to $\gamma$\nobreakdash-polysemanticity. Polysemanticity of more than two concepts can be broken down into several two-concept cases; hence, we only consider disentanglement for the two-concept case.

\subsection{Disentangling polysemantic channels}\label{subsec:disentangle}
Building on the above identification of polysemantic channels, we propose an algorithm to disentangle them in any pre-trained CNN without affecting its output or predictive performance.
Given a polysemantic channel $\channel$ and the two classes $t_1$ and $t_2$ for which the channel is relevant, we aim to disentangle the channel into two channels, each of which only activates for one of the respective class concepts.
In a nutshell, we create an artificial layer $l^\prime$ between layer $l{-}1$ and the layer $l$ containing the channel we want to disentangle.
The new layer $l^\prime$ will contain the disentangled channels, while layers $l{-}1$ and $l$ remain unchanged. %
We outline the idea in \cref{fig:neural_network}.
The modification of the network consists of three steps: \textit{(i)} construction of the new layer $l^\prime$, \textit{(ii)} connection of layer $l^\prime$ with layer $l$, and \textit{(iii)} connection of layer $l{-}1$ with the new layer $l^\prime$.

\myparagraphnospace{Construction of layer $l^\prime$} starts by copying layer $l$. We then replace channel $\channel$ in $l^\prime$ with three new channels; the first two channels will represent the two disentangled concepts while the third channel serves as a residual channel to ensure recovery of the original activations.

\myparagraphnospace{Connection of layer $l^\prime$ with $l$} consists of simple identity mappings for all corresponding channels. The first two newly introduced channels are connected with an identity mapping to the channel $\channel$ in $l$ that we aim to disentangle. The third new channel is connected to $\channel$ with a weight of ${-}1$; the reason for this follows below.

\myparagraphnospace{Connection of layer $l{-}1$ with $l^\prime$} consists of the previously existing connections from $l{-}1$ to $l$ for all but the three new channels in $l^\prime$.
For the latter, we first set weights from $l{-}1$ to the weights originally going from $l{-}1$ to $\channel$. Then, 
for each channel $i$ in $l{-}1$ we check whether it is relevant for the concept in $\class_2$ using the average relevance
\begin{equation}\label{eq:condition_handcrafted1}
\text{ARV}(l, \channel, \class_2; f)_i \cdot \rho > \text{ARV}(l, \channel, \class_1; f)_i\,,
\end{equation}
with $\rho \in [0,1]$ controlling how much larger the relevance for $\class_2$ needs to be than for $\class_1$ in order for a channel $i$ in $l{-}1$ to be considered only relevant for detecting the concept in $\class_2$. Please refer to Appendix \ref{app:experimental_setup} for an algorithm to automatically set $\rho$.
We obtain the final weight for the first of the three new channels in $l^\prime$ by setting all edges coming from $l{-}1$ to zero where  Ineq.~(\ref{eq:condition_handcrafted1}) holds. Intuitively, we have now deleted the connections to channels in $l{-}1$ that contribute to detecting concept $2$, and thus, this first new channel now mainly detects concept $1$. For the second channel, we repeat the process, now removing weights responsible for concept $1$, following an analogous procedure.
The original channel has now been ``disentangled'' into these two channels. Any features beyond these two concepts are, however, now encoded twice -- once in each of the two new channels. 
To address this issue, in the third channel, we delete all the edges coming from a channel in $l{-}1$ that is \textit{either} strongly responsible for concept $1$ \textit{or} concept $2$, \textit{i.e.}, we only keep the edges that cannot be associated exclusively to concept $1$ or $2$. As the third channel is connected to the original channel $c$ in layer $l$ with a weight of ${-}1$, it accounts for this double encoding and allows us to recover the original activation of channel $c$ in layer $l$.

\section{Experiments}
\label{sec_experiments}

The detailed experimental setup is provided in Appendix \ref{app:experimental_setup}. Briefly, we study the penultimate layer of a ResNet-50~\cite{He:2016:DRL} trained on the ImageNet dataset~\cite{Deng:2009:ILS} and select $\gamma = 0.5$. We use the ImageNet training split for disentanglement and the validation split for evaluation.

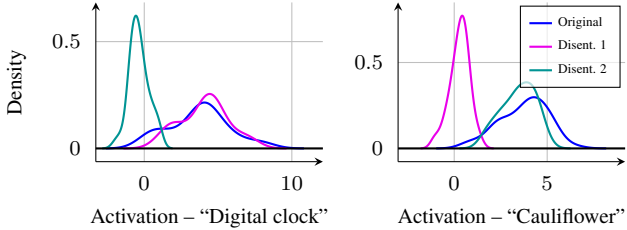
\begin{figure}[t]
    \centering
    \begin{tikzpicture}

\pgfplotsset{
compat=1.11,
legend image code/.code={
\draw[mark repeat=2,mark phase=2]
plot coordinates {
(0cm,0cm)
(0.15cm,0cm)        %
(0.3cm,0cm)         %
};%
}
}

\tikzstyle{every node}=[font=\footnotesize]
\begin{groupplot}[group style={
                      group name=myplot,
                      group size= 4 by 1},height=3.7cm,width=4.6cm,
                      axis lines = left,
                      enlarge x limits = true,
                      enlarge y limits = true,
                      tick label style={/pgf/number format/fixed},
                      xmin=-2,
                      grid=major,
                      xtick style={draw=none},
                      ytick style={draw=none},
                      legend cell align={left},
                      try min ticks=3,
                      ylabel near ticks, 
                      ylabel shift={3pt},
                      ]

    \nextgroupplot[%
    ylabel={Density}, 
    xlabel=Activation -- ``Digital clock'',
    ]
        \addplot[smooth, blue, thick] table {original_c1.dat};
        \addplot[smooth, colorConceptOne, thick] table {disentangled1_c1.dat};
        \addplot[smooth, colorConceptTwo, thick] table {disentangled2_c1.dat};
        \draw[thick] (axis cs:\pgfkeysvalueof{/pgfplots/xmin},0) -- (axis cs:\pgfkeysvalueof{/pgfplots/xmax},0);

    \nextgroupplot[%
    xlabel=Activation -- ``Cauliflower'',
    legend style={fill=white, fill opacity=0.7, draw opacity=1,text opacity=1}
    ]
        \addplot[smooth, blue, thick] table {original_c2.dat};\addlegendentry{\tiny	Original};
        \addplot[smooth, colorConceptOne, thick] table {disentangled1_c2.dat};\addlegendentry{\tiny	Disent. 1};
        \addplot[smooth, colorConceptTwo, thick] table {disentangled2_c2.dat};\addlegendentry{\tiny	Disent. 2};
        \draw[thick] (axis cs:\pgfkeysvalueof{/pgfplots/xmin},0) -- (axis cs:\pgfkeysvalueof{/pgfplots/xmax},0);

\end{groupplot}

\end{tikzpicture}
    \vspace{-2em}
    \caption{\textit{Density plot of the activations for the original channel $\#1660$ and the corresponding disentangled channels for images from the classes ``digital clock'' and ``cauliflower''.} In both plots, one disentangled channel mimics the activation of the original channel $\#1660$ while the other disentangled channel is mostly inactive, indicating that the disentanglement was successful.}
    \label{fig:density}
    \vspace{-0.5em}
\end{figure}

\myparagraph{Qualitative analysis. }
As an illustrative example, we found that channel $\#1660$ is relevant for the classes ``digital clock'' and ``cauliflower''. Their ARVs have a relatively low cosine similarity of $0.47$ (see \cref{def:superposition}), thus the channel is affected by $\gamma$\nobreakdash-polysemanticity. %
We disentangle channel $\#1660$ to obtain two channels that detect the corresponding ``digital clock'' and ``cauliflower'' concepts, respectively (see \cref{subsec:disentangle}).
To verify the disentanglement, we feed evaluation images showing one of the two classes of interest through the original and disentangled model and plot the density function (kernel density estimate) for the activation of the original channel $\#1660$ and the two disentangled channels in \cref{fig:density}. %
Remarkably, the first disentangled channel mimics the activation of the original channel for the ``digital clock'' images while barely responding to ``cauliflower'' images, with the converse holding for the second channel, suggesting successful disentanglement. %

To better understand the concepts a channel encodes in an image classification model, we provide images for which the channel activates the most and additionally use feature visualization by optimization~\cite{Erhan:2009:VHL}.
However, visualizing a polysemantic channel presents two challenges: The visualization may highlight only one concept, partially obscuring the channel's function, or it may capture both concepts together, making the explanation less interpretable~\cite{Olah:2017:FV}.
Our disentangled representation of a channel alleviates this problem, which we exemplify by visualizing the concepts activating channel $\#1660$ and the two disentangled channels, respectively, in \cref{fig:visualization_handcrafted}. To visualize a feature by optimization, we use MACO~\cite{Fel:2023:UFV}. 
For the original channel, we see that it is most strongly activated for images from ``digital clock'' and ``cauliflower'' alike, and even in the MACO visualization, we see that both cauliflower patterns and digital clock parts are visible. The disentangled channels only activate for images of one of the two classes, and in the MACO visualization, the respective concepts appear more isolated for the disentangled channels. We also observe that some cauliflower patterns are still lightly visible in the MACO visualization for the first disentangled channel. This might be due to the two disentangled channels still sharing various edges that have not been found to be particularly relevant for the concept in class 1, respectively, class 2, according to Ineq.~(\ref{eq:condition_handcrafted1}).%

\myparagraph{Quantitative analysis.}
We next consider all polysemantic channels in the penultimate layer --~affected by $\gamma$\nobreakdash-polysemanticity with $\gamma=0.5$~-- for a quantitative analysis. %
For each such channel, we measure the mean relative activation after disentanglement --~\ie, the activation divided by the activation of the original (entangled) channel~-- of the corresponding disentangled and residual channels for evaluation images of the two classes of interest in \cref{tab:disentanglement}. %
Confirming insights from the qualitative analysis, %
the first new channel almost exactly mimics the activation of the original channel ($114\%$) for concept 1 images while only marginally ($-9\%$) activating for concept 2 images. We see the opposite trend for the second new channel with $-21\%$ and $125\%$ activation for concept 1 and concept 2, respectively. %
Moreover, the residual channel's relative activation is very close to zero, suggesting it is only responsible for a minuscule amount of information. Thus, it does not necessarily need to be explained explicitly.
These promising results indicate that our proposed algorithm for disentangling polysemantic channels is effective.

\begin{table}
\centering
\small
\begin{tabularx}{\linewidth}{@{}Xccc@{}}
\toprule

Concept & Disentangled 1 & Disentangled 2 &
           Residual \\ 
        \midrule
           Concept 1 & $114$ $\uparrow$ & $-9$ $\downarrow$ & $0$ $\downarrow$ \\ 

Concept 2 & $-21$ $\downarrow$ & $125$ $\uparrow$ & $0$ $\downarrow$\\

\bottomrule
\end{tabularx}
\vspace{-0.5em}
\caption{\textit{Quantitative evaluation.} We report the activation for the disentangled and residual channel relative to the activation of the original channel (in $\%$) for images from the first (Concept 1) and second class (Concept 2).}
\label{tab:disentanglement}
\vspace{-0.5em}
\end{table}

\section{Conclusion \& Limitations}
\label{sec:conclusion}

We introduce a new mechanistic interpretability tool that identifies and disentangles channels affected by $\gamma$\nobreakdash-polysemanticity into multiple channels, each only activating for one of the concepts.
While not directly applicable to other forms of polysemanticity, our formal definition of $\gamma$\nobreakdash-polysemanticity
allows us to assess the effectiveness of our approach, both qualitatively and quantitatively. %
Currently only applied to the common form of disentanglement of \textit{two} concepts, we anticipate that our method is easily extendable to more classes, \textit{e.g.}, through a recursive application. %
In this initial work, we focused on a specific set of hyperparameters that gave reasonable results and aligned with previous knowledge (\textit{e.g.}, association of ARV similarity with WordNet similarity; see Appendix \ref{app:experimental_setup}), yielding a proof of concept for our general approach.
Future work could further explore the impact of these hyperparameters and potential methods for setting them automatically.
Additionally, our current formulation of polysemanticity is based on class labels, thus, our proposed disentanglement may work better in later layers with more semantically meaningful concepts.  
Although preliminary, we are confident that our work will be a valuable extension of the current landscape of mechanistic interpretability.

\newpage

\paragraph{Acknowledgments.}
This project has received funding from the
European Research Council (ERC) under the European Union’s Horizon 2020 research and innovation programme (grant agreement No.~866008).
The project has also been
funded by the Deutsche Forschungsgemeinschaft (DFG, German Research Foundation) --~project number 529680848. Further, the project has been supported by the State of Hesse, through the cluster projects ``The Third Wave of Artificial
Intelligence (3AI)'' and ``The Adaptive Mind (TAM),'' and by the Hessian Ministry of Higher Education, Research, Science and the Arts
and its LOEWE research priority program ``WhiteBox'' under grant LOEWE/2/13/519/03/06.001(0010)/77. 

{
    \small
    \bibliographystyle{ieeenat_fullname}
    \bibliography{bibtex/short,bibtex/external,bibtex/local}
}

\newpage
\appendix
\section{Experimental Details}\label{app:experimental_setup}
\newcommand{\argmax}{\operatornamewithlimits{arg\,max}}
For our experiments, we use the ImageNet dataset~\cite{Deng:2009:ILS} and study the popular ResNet-50 architecture~\cite{He:2016:DRL}. %
We use pre-trained model weights from PyTorch~\cite{Paszke:2017:ADP}. Our layer of interest $l$ is the last convolutional layer -- earlier layers could have been used as well; however, we chose the last one for simplicity and because we suspect that later layers encoding semantically more meaningful concepts are easier to disentangle with our proposed approach. We consider a channel $\channel$ as relevant for class $\class$ if the condition in Ineq.~(\ref{eq:relevant_channel}) with $\threshold = 0.03$ and $p = 0.75$ is satisfied. We consider a channel affected by polysemanticity if the cosine similarity in \cref{def:superposition} is below $\thresholdARV=0.5$. 
The values of $\threshold$, $\thresholdARV$, and $p$ control at what point a channel is considered relevant for a class and at what point it is affected by polysemanticity. Therefore, they are highly application-specific and can be selected more or less conservatively depending on whether the application requires higher recall or higher precision (in this work, $\threshold$, $\thresholdARV$, and $p$ are handpicked). 

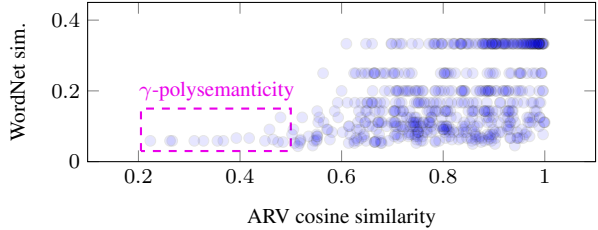
\begin{figure}[t]
\centering
\begin{tikzpicture}
    \footnotesize
    \begin{axis}[xmin=0.1,xmax=1.1,
        ymin=0.00,ymax=0.45,
        width=1\linewidth, %
        height=3.7cm,
        ylabel style={yshift=-0.3cm},
        ylabel={WordNet sim.}, 
        xlabel={ARV cosine similarity},
        try min ticks=2,
        ]

    \addplot[only marks,mark=*, mark options={fill=blue,scale=1,fill opacity=0.1, draw opacity=0.1,}] table {cossim_wordnet.dat};

    \draw[dashed, colorConceptOne, thick] (10.5,30) rectangle (40,150) node[anchor=south, xshift=-28] {\footnotesize $\gamma$-polysemanticity};
    
    \end{axis}
\end{tikzpicture}
\vspace{-0.8em}
\caption{\textit{WordNet similarity of the two classes for which a channel is relevant over their ARV cosine similarity %
(see Definition \ref{def:superposition}).} %
}
\vspace{-1.0em}
\label{fig:scatter_wordnet}
\end{figure}

More specifically, %
the hyperparameter $\gamma$ controls the maximum $\text{ARV}$ cosine similarity for which a channel relevant for two classes is considered to be polysemantic.
We aim to choose $\gamma$ to include as many channels as possible while maximizing the chance that the included channels are truly polysemantic and not activating for a shared concept, such as for different dog species.
To find an appropriate value for $\gamma$, we take all class pairs for which the same channel $c$ is relevant (see Ineq.~(\ref{eq:relevant_channel})) and plot the WordNet~\cite{Deselaers:2011:VSS} path similarity for the two class labels over their $\text{ARV}$ cosine similarity in \cref{fig:scatter_wordnet}. Intuitively, similar classes that share concepts, such as different dog species, have a high WordNet similarity while semantically different classes have a low similarity. Thus, we can use the WordNet similarity as a proxy for semantic similarity, which we use as a proxy for visual similarity~\cite{Deselaers:2011:VSS}.
For $\gamma=0.5$, almost all the relevant class pairs have a very low WordNet similarity, and thus, we continue our analysis with class pairs that have an ARV cosine similarity below that value.

The factor $\rho$ controlling when a channel $i$ in $l{-}1$ is mainly relevant for the concept of $\class_2$, respectively $\class_1$, in Ineq.~(\ref{eq:condition_handcrafted1}) is selected automatically. To this end, we take all training images from $\class_1$ and $\class_2$ and measure the activations of the disentangled channels $c^\prime_1$ and $c^\prime_2$ for different $\rho$ to maximize
\begin{equation}
\begin{split}\argmax_\rho
    &\frac{1}{|X^{(\class_1)}|} \sum_{x \in X^{(\class_1)}} \frac{\operatorname{sum} \left( |f_{l^\prime}(x)_{c^\prime_1}| \right)}{\operatorname{sum} \left( |f_{l^\prime}(x)_{c^\prime_2}|\right)} \\ +  &\frac{1}{|X^{(\class_2)}|} \sum_{x \in X^{(\class_2)}} \frac{\operatorname{sum} \left( |f_{l^\prime}(x)_{c^\prime_2}| \right)}{\operatorname{sum} \left( |f_{l^\prime}(x)_{c^\prime_1}|\right)}.
    \end{split}
\end{equation}

\paragraph{Qualitative analysis.} To visualize the highest activating image patches in \cref{fig:visualization_handcrafted} from two classes, we select the 16 highest activating images and extract the highest activating patches. We then choose six patches such that they are equally distributed among the two classes (if there are images from both classes within the 16 patches). This allows us to consider the top 16 patches, without overfilling the plot.

\paragraph{Quantitative analysis.} For our quantitative analysis, the relative activation is meaningless for images in which the concept encoded in the channel of interest is absent. Thus, we only include images where the channel of interest has a relative attribution above $\threshold$ (see Ineq.~(\ref{eq:relevant_channel})). %
With the chosen hyperparameters, %
this condition applies to at least $75\%$ of the images, representing a sufficiently large number to draw meaningful conclusions.

\end{document}